\documentclass{article}

\usepackage[final]{neurips_2019}

\usepackage[utf8]{inputenc} 
\usepackage[T1]{fontenc}    
\PassOptionsToPackage{hyphens}{url}\usepackage{hyperref}
\usepackage{url}            
\usepackage{booktabs}       
\usepackage{amsfonts}       
\usepackage{nicefrac}       
\usepackage{microtype}      

\usepackage{graphicx}
\usepackage{subfigure}
\usepackage{amsmath}
\usepackage{amssymb}
\usepackage{caption}
\usepackage{paralist}
\usepackage{xspace}
\usepackage{color}
\usepackage{wrapfig}
\usepackage{array}
\usepackage{float}
\usepackage{todonotes}
\setcitestyle{square,numbers}

\usepackage{algorithm}
\usepackage[noend]{algpseudocode}
\usepackage{xcolor}
\hypersetup{
    colorlinks,
    linkcolor={red!50!black},
    citecolor={blue!50!black},
    urlcolor={blue!70!black}
}

\usepackage{comment}

\newcolumntype{C}{>{\centering\arraybackslash}m{0.3\textwidth}}

\newcommand{\imgep}{\textsc{imgep}\xspace}
\newcommand{\uvfa}{\textsc{uvfa}\xspace}
\newcommand{\her}{\textsc{her}\xspace}

\newcommand{\ddpg}{\textsc{ddpg}\xspace}
\newcommand{\curious}{\textsc{curious}\xspace}
\newcommand{\unicorn}{\textsc{unicorn}\xspace}

\usepackage[toc,page]{appendix}

\definecolor{myred}{rgb}{0.8,0,0}

\title{Language Grounding through Social Interactions and Curiosity-Driven Multi-Goal Learning}

\author{
    Nicolas Lair\thanks{joint first authors}\\
    Cloud Temple, INSERM \\
    France\\
    \texttt{nicolas.lair@inserm.fr}\\
   \And
    C\'edric Colas\footnotemark[1] \\
    INRIA\\
    France\\
    \texttt{cedric.colas@inria.fr}  
  \And
    R\'emy Portelas \\
    INRIA\\
    France\\
    \texttt{remy.portelas@inria.fr}\\
  \AND
    Jean-Michel Dussoux \\
    Cloud Temple\\
    France\\
    \texttt{jean-michel.dussoux@cloud-temple.com}\\
  \And
    Peter Ford Dominey \\
    INSERM\\
    France\\
    \texttt{peter.dominey@inserm.fr}\\
  \And
    Pierre-Yves Oudeyer \\
    INRIA\\
    France\\
    \texttt{pierre-yves.oudeyer@inria.fr}\\
  }
\begin{document}

\maketitle

\begin{abstract}
Autonomous reinforcement learning agents, like children, do not have access to predefined goals and reward functions. They must discover potential goals, learn their own reward functions and engage in their own learning trajectory. Children, however, benefit from exposure to language, helping to organize and mediate their thought. We propose LE2 (Language Enhanced Exploration), a learning algorithm leveraging intrinsic motivations and natural language (NL) interactions with a descriptive social partner (SP). Using NL descriptions from the SP, it can learn an NL-conditioned reward function to formulate goals for intrinsically motivated goal exploration and learn a goal-conditioned policy. By exploring, collecting descriptions from the SP and jointly learning the reward function and the policy, the agent grounds NL descriptions into real behavioral goals. From simple goals discovered early to more complex goals discovered by experimenting on simpler ones, our agent autonomously builds its own behavioral repertoire. This naturally occurring curriculum is supplemented by an active learning curriculum resulting from the agent's intrinsic motivations. Experiments are presented with a simulated robotic arm that interacts with several objects including tools.
\end{abstract}

\section{Introduction}
Reinforcement Learning (RL) algorithms traditionally focus on learning to solve specific and well-defined tasks. This implies a prior definition of the task space and the associated reward functions. Designing reward functions is not always straightforward, and this needs to be done for every new task, making the definition of RL problems cumbersome.

This is quite different from the way a child learns. A child has to learn at the same time what the goals are, how to perform them and how to evaluate his performance. Interactions with adults are then critical for the child. It is through the combination of social interactions and real world experimentation that a child learns how to ground natural language (NL) in his experience of the world, and use it as a tool to explore further \cite{Vygotskii1978}.

Based on these observations, we propose an RL-based algorithm that learns to master a series of goals through autonomous exploration and interactions in natural language with an informed social partner. The agent freely explores its environment, has full control over its learning curriculum but has no access to any external reward function or any predefined goal space. It leverages the knowledge of a social partner who describes in NL sentences what has been achieved by the agent in an episode. The agent turns these descriptions into targetable goals, effectively grounding natural language in concrete behaviors through interactions with the environment. Leveraging these descriptions, the agent learns its own reward function and uses it to train on already discovered goals. Training on simple goals and exploring around them enables the discovery of more complex goals, which generates a natural curriculum. In addition, our agent sets its own goals and uses intrinsic motivations to guide its own learning trajectory towards the collection of high-quality data and towards the maximization of learning progress. 

\paragraph{Related work -}The idea that language understanding is grounded in one's experience of the world and should not be separated from the perceptual and motor systems \cite{Glenberg2002, Zwaan05} has a long history in Cognitive Sciences. The transposition of the idea of embodied language into intelligent systems \cite{steels2006semiotic} was then applied to human-machine interaction \cite{Dominey2005, Madden2010} and to semantic representations \cite{Silberer2012}. Recently, the \textit{BabyAI} framework was proposed as a platform to pursue Deep RL research in grounded language \cite{chevalier-boisvert2018babyai}.

In the domain of RL, the idea that natural language could provide a guidance for an algorithm has recently gained in popularity. In their review on \textit{RL algorithms informed by natural language}, \citet{Luketina2019} distinguish between \textit{language-conditional} problems where language is required to solve the task and \textit{language-assisted} problems where language is a supplementary help. In the first category, most works propose instruction-following agents \cite{bahdanau2018learning, Jiang2019, Goyal2019, Branavan2010, Chen2011}. Although our work might fall in the language-conditional category, our system is never given any instruction. It builds its own learning curriculum depending on its experience and learns its own reward function. \citet{bahdanau2018learning} also learns a reward function jointly with the action policy but does so using a dataset of expert data whereas our agent uses trajectories collected through its own exploration. Lastly, \citet{fu2018from} uses an inverse RL approach to learn a reward function, but requires known environment dynamics.

\textit{Curriculum learning} refers to the organization of a learning trajectory into distinct training phases, usually starting simple and gradually complexifying the task \citep{elman,krueger,bengiocl}. In the multi-task RL framework, curriculum learning organizes the sequence of training tasks presented to the agent, so as to maximize its performance on a set of target tasks, leveraging transfer learning \citep{taylortransfer}. In developmental robotics finally, learning progress is often used as an intrinsic motivation to guide the learning trajectory. The agent tracks its own progress and can guide the selection of goals to target, thus actively adapting its curriculum depending on its performance \cite{saggriac, modularforestier, colas2019curious}.


\section{Methods}
\subsection{Problem Definition}
In the present paper, the agent evolves in the environment without any prior on the set of possible goals and without external reward functions. The only source of guidance comes from a social partner (SP), an entity providing natural language descriptions of the behavior of the agent. In practice, the social partner is a function using a hard-coded list of potential goals. Given a trajectory of the agent, it returns the list of descriptions corresponding to the goals achieved in that trajectory expressed in natural language (e.g. \textit{Grasp the magnetic stick}). In this setting, the objective of the agent is to maximize its average competence on the set of goals. However, the agent has to discover achievable goals through its own exploration, while learning how to reach those already discovered. A successful agent is an agent that discovers and masters the maximum number of goals. \textbf{In summary, this paper defines a novel use of language grounding where the goals and reward functions are learned through exploration and grounded by descriptions from the SP}.


\subsection{An Autonomous Agent in an Unknown Multi-Goal Setting}
\label{sec:algo_description}

\begin{figure}[ht]
\centering
    \includegraphics[width=0.65\columnwidth]{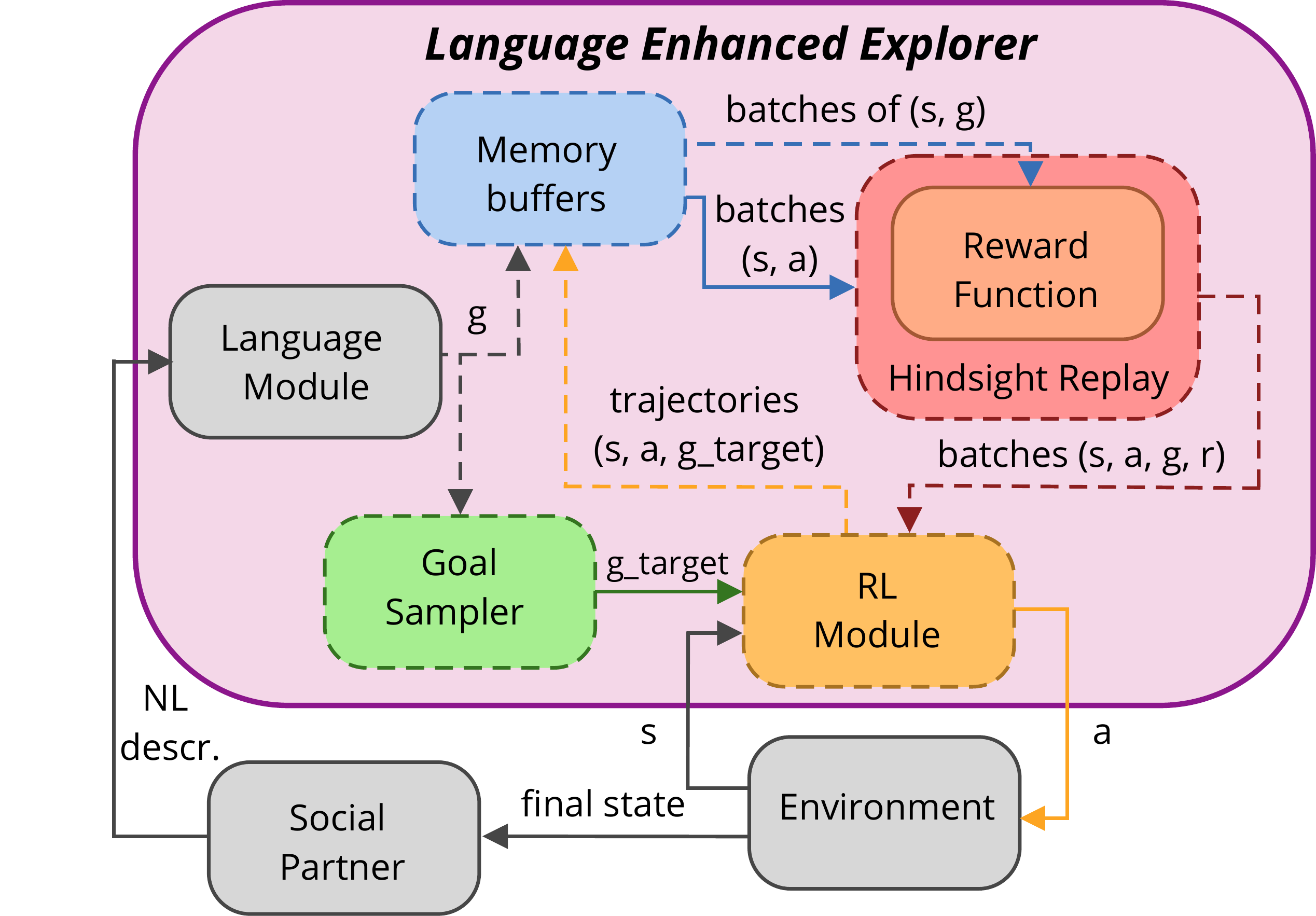}
     \caption{\textbf{The \emph{Language Enhanced Exploration} algorithm.} Colored dashed circled boxes represent modules that are updated by the LE2 algorithm. Grey boxes represent non-updated modules, some of which are external to the agent (social partner, environment).}  
\label{fig:le2}
\end{figure}

\paragraph{Architecture -}Figure \ref{fig:le2} represents the Language Enhanced Exploration algorithm. The learning agent samples a goal to target among previously discovered goals (or noise if there is none, green) and interacts interacts with its environment (yellow). While the trajectory collected by the agent is used to update its memory (blue), the social partner receives the final state and provides corresponding descriptions in natural language. The agent encodes these NL descriptions into goal encodings $g$ (language module) and stores them in memory. Now that the agent collected information in memory, the algorithm needs to update three other modules: the goal sampler, the reward function and the RL module (policy and critic). The goal sampler is updated using goal encodings decoded from the social partner's feedbacks. The reward function uses pairs of final states and associated goal encodings. Finally, to update the RL module, the algorithm samples a batch of station-action transitions from the memory and uses hindsight replay to figure out which goals to train on. The hindsight module uses the reward function to imagine what would be the reward if a transition had been collected while aiming at a particular goal. Using this process, it finds candidate goals leading to rewards, and can regulate the ratio of transitions with positive versus negative rewards. This process augments the batch of state-action pairs with goal encodings and reward information. \ddpg then uses this augmented batch to update the policy and critic.

\paragraph{From Language to Goal Vectors -}The agent must build a goal space from its experience in the environment and the descriptions provided in natural language by the SP. Each description is associated to a potential goal for the agent. This goal is represented by a fixed-size vector obtained by averaging the embeddings of each word contained in the SP description. Word embeddings are obtained using a pre-trained GLOVE model \cite{pennington2014glove}.


\paragraph{Multi-Goal Reinforcement Learning Agent -}Our agent is controlled by a goal-conditioned policy $\pi$ \cite{schaul2015universal}. This policy takes as input the state of the world $s_t$ at time $t$ and its current goal $g$ (encoded from a SP description) and returns the next continuous action vector $a$. The current state is the concatenation of the current observations and the difference between the current observations and the initial observations $s_t = o_t - \Delta o_t = o_t - (o_t - o_0)$  We consider fixed-length episodes, where the agent sets its own goal for the whole episode. Our agent belongs to the framework of \imgep \cite{forestier2017intrinsically} and is based on the \curious \cite{colas2019curious} algorithm. In the same spirit it sets its own goals and uses intrinsic motivations to guide its learning trajectory.

Our agent is based on a goal-conditioned actor-critic learning algorithm (\ddpg) augmented by the idea of Hindsight Experience Replay (\her) \cite{andrychowicz2017hindsight}. As in \her, our agent can use transitions collected while aiming at a particular goal $g_i$ to learn about any goal $g_j$ by \textit{replay}. In practice, the original goal $g_i$ contained in a transition $([s_t, g_i], a_t)$ can be substituted by any other goal $g_j$ the agent might want to learn about. In simpler words, the agent \textit{replays} the memory of a trajectory (e.g. when trying to reach object 1) by \textit{pretending} it was targeting a different goal (e.g. reaching object 2). Given this replayed transition, the agent can compute the associated reward $r_t$ given its internal reward function. We explain below how the agent learns this internal reward function.


\paragraph{Curriculum and Intrinsic Motivations -}The agent discovers new goals only as it performs them for the first time through its own exploration. Note that this results in a naturally-occurring curriculum. The agent has more chance to solve the simplest goals by chance, thus discovers them first. Harder goals can hardly be solved by chance at first, but might be discovered as the agent explores while learning to master simpler ones. As it learns, the agent stores newly discovered goals in a list of targetable goals. In addition to this implicit curriculum, we endow the agent with two types of intrinsic motivations, enabling it to shape its learning trajectory (active curriculum).

The first intrinsic motivation biases the agent to target goals depending on their associated probabilities to generate \textit{high-quality trajectories}. High-quality trajectories are trajectories where the agent collects descriptions from the social partner for goals that are rarely reached. The expected quality of a trajectory given the current policy $\pi$ and the targeted goal $g_t$ is defined by $\mathbb{E}(quality \mid g_t, \pi) = \sum_i p(d_i \mid g_t, \pi) \times rarity(g_i)$ where $p(d_i^+ \mid g_t, \pi)$ is the probability to obtain a description $d_i$ (corresponding to goal $g_i$) when targeting $g_t$ with policy $\pi$ and the \textit{rarity score} $rarity(g_i)$  is the inverse of the number of descriptions collected for goal $g_i$.

The second intrinsic motivation biases the selection of goals to substitute in transitions used for updating the policy and critic (i.e. goals to learn about). For each transition sampled from the replay buffer, the agent uses its internal reward function to determine which goals are considered solved in that transition. Given this set of goals, the agent prioritizes goals with higher absolute learning progress (ALP), as in \citet{colas2019curious}. Further details on the intrinsic motivations in Appendix \ref{app:intrinsic}.



\paragraph{Learning a Goal-Conditioned Reward Function -}The policy $\pi$ and the internal reward function $\mathcal{R}_i$ are learned in parallel. The agent collects transitions as it experiences the environment and uses the descriptions provided by the SP to learn $\mathcal{R}_i$. At any time during learning, the agent has discovered the set of goals $\mathcal{G}_d$. For each trajectory SP provides descriptions about all goals achieved by the agent $\mathcal{G}_a$. The agent can thus infer the set of goals $\mathcal{G}_m$ that were not achieved in this trajectory by taking the complement $\mathcal{G}_m~=~\mathcal{G}_d~-~\mathcal{G}_a$. Given environment interactions and feedback from SP, the agent builds a dataset with $[s_T, g_i, r_i^+] \xspace \forall g_i \in \mathcal{G}_a$ and $[s_T, g_i, r_i^-] \xspace \forall g_i \in \mathcal{G}_m$, $T$ being the final time step.

Learning the reward function $\mathcal{R}_i(s, \Delta s, g): \mathcal{S} \times \mathcal{G} \to \{ 0, 1\}$ is framed as a goal-conditioned binary classification problem, where $\mathcal{S}, \mathcal{G}$ are the state space and goal space respectively. $\mathcal{R}_i$ is periodically updated using a random forest classification algorithm, as it is fast to train and predict. The agent is now able to estimate whether a given state $s_t$ corresponds to a reward for each of the known goals. Further details can be found in Appendix \ref{SPfeedback}.

\section{Experiment and Results}

\paragraph{Agent Evaluation -} To evaluate our agent, we present \textit{ArmToolsToys}, an environment adapted from \cite{forestier2016modular} consisting in a robotic arm that learn to use tools to grasp some objects. The SP defines $51$ potential goals in this environment (see Appendix \ref{envdetails}). We evaluate our agents offline every 600 training episodes on the full set of $51$ goals (undiscovered goals included) using the true reward function. We monitor the quality of the learned reward function using the F1-score, comparing the prediction of the learned reward function to the ground truth, using transitions recently collected by the agent (but not used to train the reward function) and averaged over discovered goals.

\begin{figure*}[ht!]
  \centering
  \subfigure[\label{fig:perf}]{\includegraphics[width=0.45\textwidth]{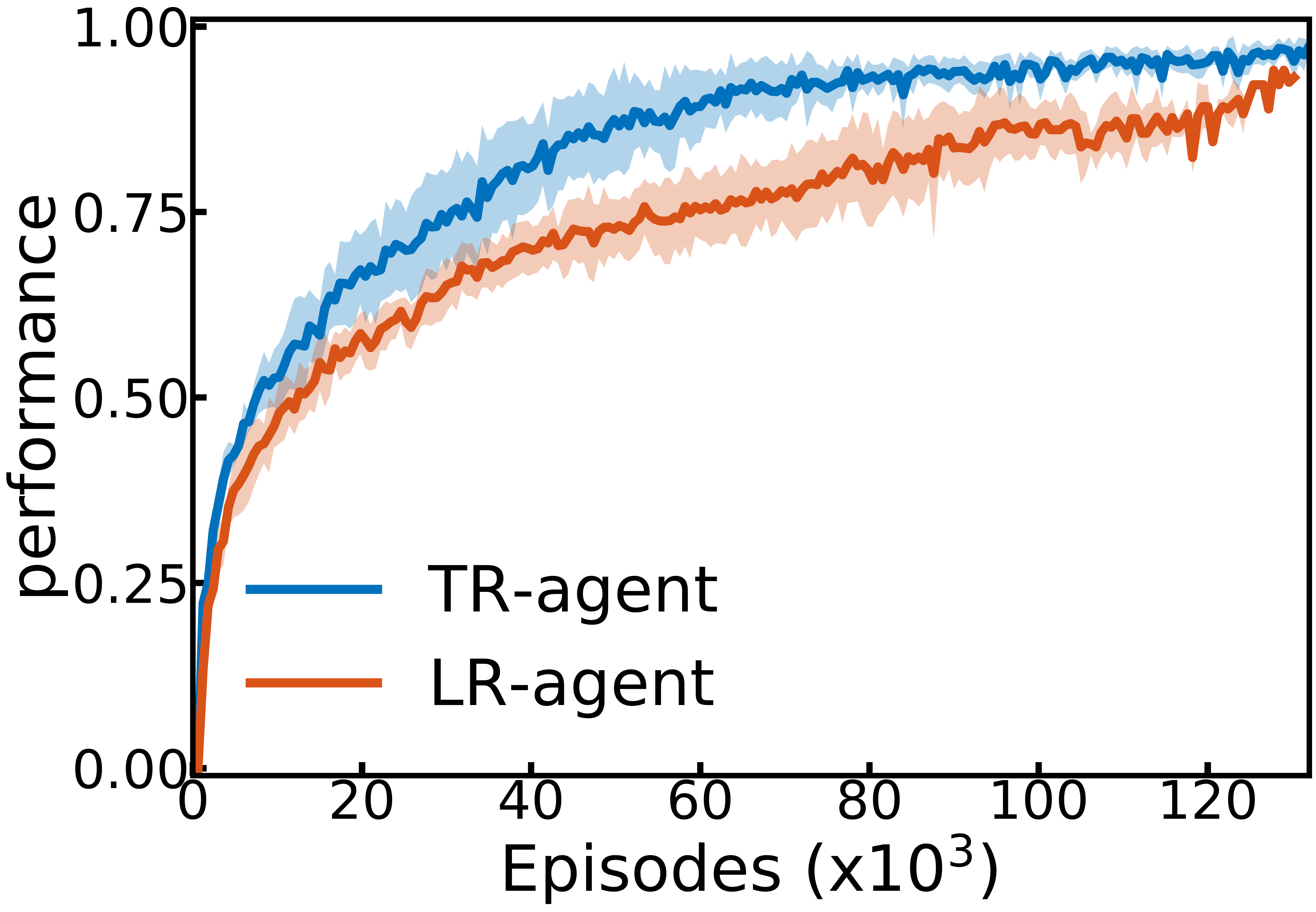}}
  \subfigure[\label{fig:f1score}]{\includegraphics[width=0.45\textwidth]{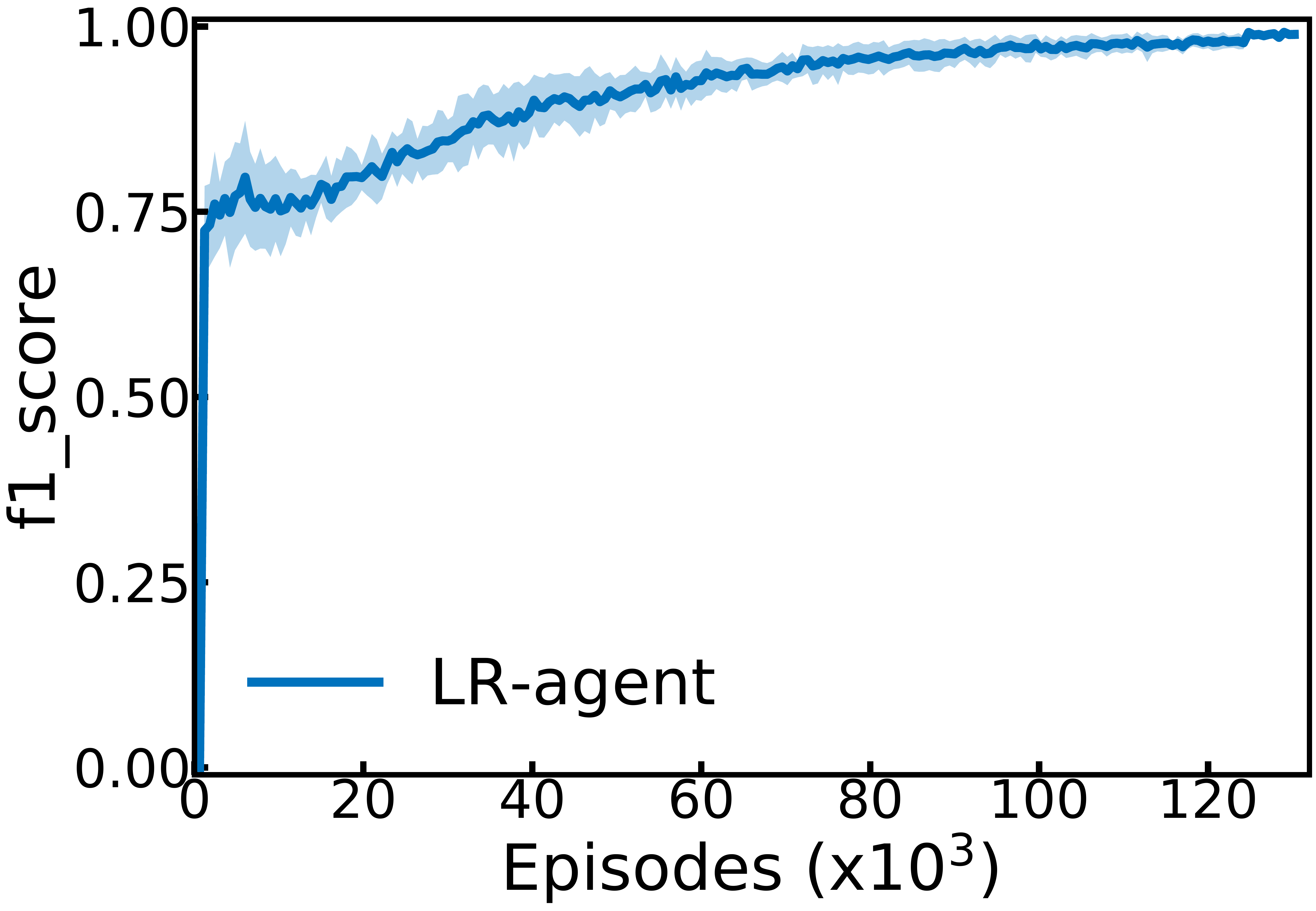}}
  \subfigure[\label{fig:goalproba}]{\includegraphics[width=0.45\textwidth]{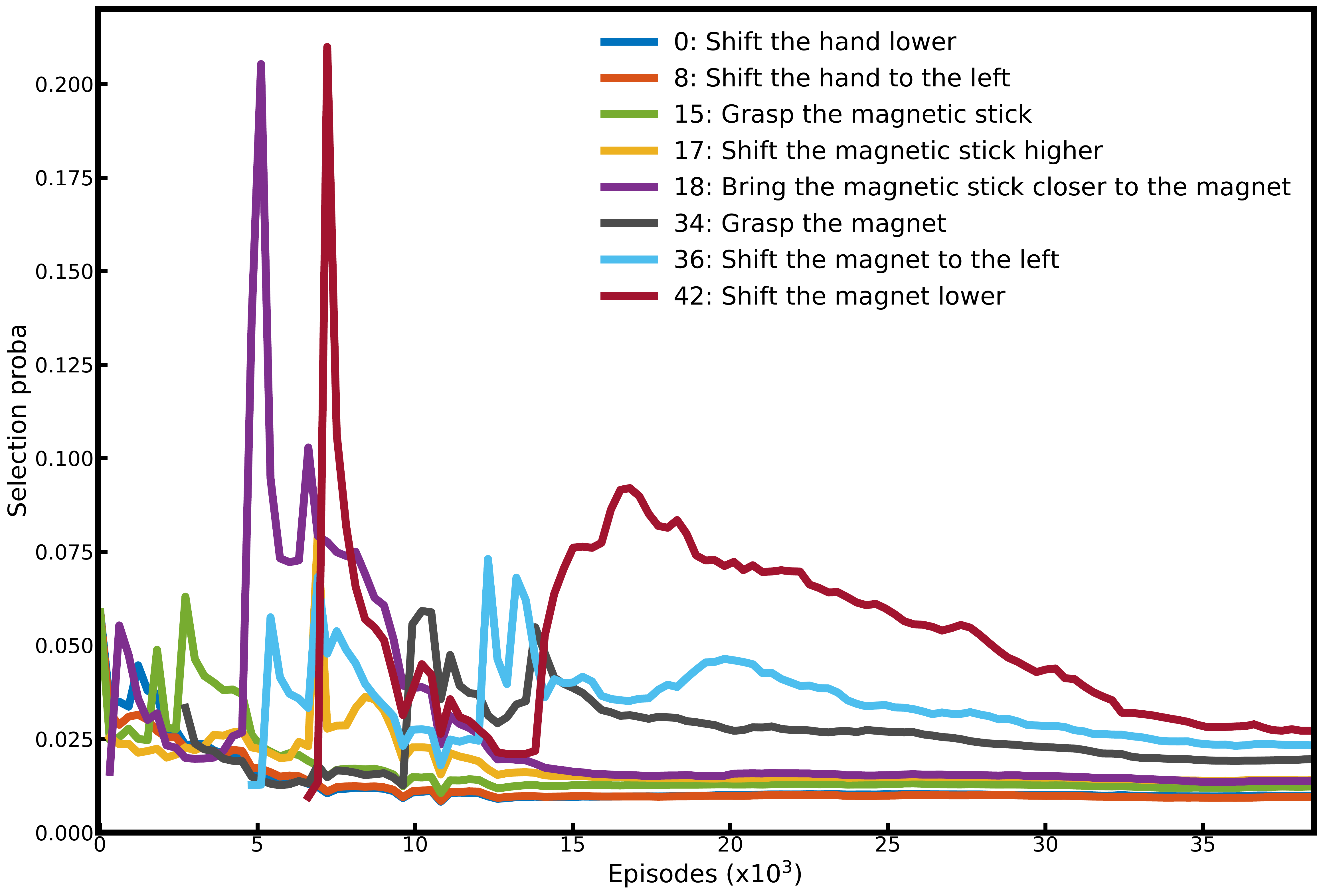}}
  \subfigure[\label{fig:parallellearning}]{\includegraphics[width=0.45\textwidth]{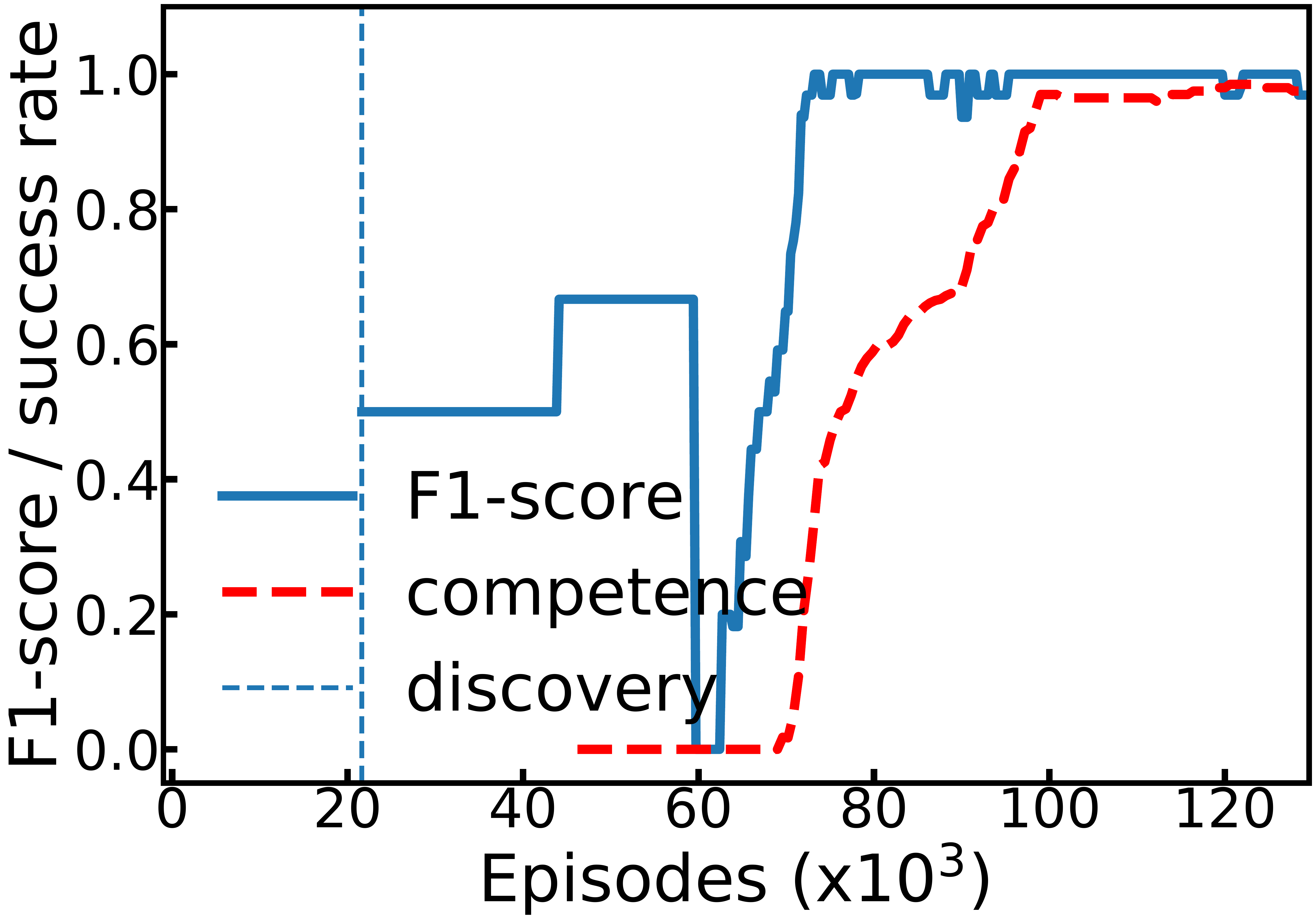}}
  \caption{\textbf{Agent evaluation.} a: Success rates averaged over all 51 goals (mean, std). b: F1-score averaged over the discovered goals (mean, std). c: Goal selection probability during a run. d: Metrics (reward model \& policy) for the goal \textit{move the magnet to the bottom right area} during a run.}
\end{figure*}

\paragraph{Learning a reward function and a policy -}Figure \ref{fig:perf} shows the evolution of the average success rate computed over all $51$ goals. The agent using the true reward function (TR-agent) naturally performs the best (max $98\%$). The agent learning its own reward function from SP feedbacks (LR-agent) is as efficient as TR-agent to learn the first half of the tasks (the simpler ones) and then gets slower but still achieves to  master almost all the tasks. Figure \ref{fig:f1score} measures the quality (F1-score) of the predictions given by the learned reward function over time on the discovered goals. Because the reward function does not produce enough accurate predictions at the beginning, the LR-agent is slower than TR-agent. However, we see that the joint learning is successful as both the reward function and the policy achieved almost perfect score.

\paragraph{Curriculum -}
Figure \ref{fig:goalproba} represents the target sampling probabilities for a subset of the available goals across time. High probability means a high interest of the agent for the goal. This intrinsic motivation results in an automatic curriculum where the agent's interest shifts from simpler to harder goals as it collects rewards and discover new goals. For a given goal, Figure \ref{fig:parallellearning} shows the time of discovery and the learning curves of the reward model and the policy. The progress of the reward model to evaluate the task are directly followed by the progress of the policy to realize it.



\section{Discussion and Conclusion}
In this work, we demonstrate that through interactions in natural language with a descriptive SP, an agent can learn to ground descriptive sentences into behavioral goals through the joint learning of a goal-conditioned reward function and policy. The resulting agent displays a high level of autonomy as they do not require prior knowledge of the goal space or the reward functions. Although our SP possesses that prior knowledge, it could theoretically be replaced by a human subject with intuitive knowledge of achievable goals. This would remove the need to hardcode a list of goals and reward functions. Our agents progressively discover new goals while trying to achieve known ones. They build and extend their own behavioral repertoire as they explore their environment and interact with the SP. 

\paragraph{Future Work -}Future work will extend this algorithm to deal with more realistic feedback, allowing the social partner to be absent or give partial descriptions of trajectories. The mapping from language to goals is currently implemented by a pre-trained language model, but could be specifically trained for the task in an end-to-end manner. This would more efficiently ground the learning of language in the experience of the agent, enabling some generalization across similar descriptions (same description said in other words, transposition of a description from an object to another) and more efficient transfer learning between semantically close goals.


\subsection*{Acknowledgments}
Nicolas Lair is supported by ANRT/CIFRE contract No. 151575A20 from Cloud Temple. C\'edric Colas is partly funded by the French Minist\`ere des Arm\'ees - Direction G\'en\'erale de l’Armement.

\bibliography{biblio}
\bibliographystyle{unsrtnat}

\newpage
\begin{appendices}

\section{Learning Algorithm}
Our algorithm is based on a parallel implementation of \ddpg + \her from OpenAI baselines \cite{lillicrap2015continuous, andrychowicz2017hindsight}.\footnote{\url{https://github.com/openai/baselines}} 6 independent workers interact with the environment and collect transitions that are stored in 6 independent replay buffers. Each worker then performs updates by sampling transitions from its replay buffer using \ddpg. These updates are then summed over the set of workers to update the policy that is then dispatched to all the workers. \ddpg is an actor-critic algorithm where the actor implements the policy and the critic approximates the optimal action-value function. In our case, the policy and the critic are \uvfa, which means they are conditioned on a current goal. Here, the goal is a goal-encoding corresponding to a description in natural language (see subsection \ref{sec:algo_description} in the article). The input of the policy is the concatenation of the current state of the world $s_t$, the difference between the current state and the initial state $\Delta s_t=s_t-s_0$ and the current goal $g$. The input of the critic is the concatenation of the policy's input and the action $a_t$. We use the default parameters of the OpenAI baselines \her implementation.

\section{Intrinsic Motivations}
\label{app:intrinsic}
\paragraph{Sampling relevant transitions from memory -} Because some goals are easy to achieve while others are much more difficult, the replay buffers usually contain mainly transitions that satisfy easy goals, and only a few satisfying harder goals. To increase learning efficiency, each worker maintains the main replay buffer containing all transitions and goal-specific indices buffers that point towards episodes where the goal has been reached. By sampling uniformly from the set of goal-specific buffers before sampling the exact index of a transition, we maintain a balance between transitions satisfying easy and hard goals.

\paragraph{Intrinsic motivations for target goal selection -} We introduce two intrinsic motivations, one for the selection of the next goal to target, the other for the selection of goals to replay. Selecting its own goals is a property of autonomous agent from the \imgep framework. The choice of the goal to target influences the trajectories, hence the quality of the data collection. For a given trajectory resulting in positive rewards for goals $(g_i)_{i\in \mathcal{G}_a}$, we define the quality of the trajectory as $q=\sum_{i\in I}rarity(g_i)$ where the \textit{rarity score} $rarity(g_i)$ is the inverse of the number of rewards collected for goal $g_i$. Maximizing the quality of data collection can therefore be formulated as a bandit problem where the bandits are the potential target goals and the value to maximize is the expected quality of the next trajectory given the target goal and the current policy: 

$$
value = \mathbb{E}(quality|g_t, \pi) = \sum_i p(r_i^+ \mid g_t, \pi) \times rarity(g_i),
$$
where $p(r_i^+ \mid g_t, \pi)$ is the probability to obtain a positive rewards for goal $g_i$ when $g_t$ is targeted given the current policy $\pi$. This can be estimated by the corresponding empirical frequencies $freq(r_i^+ \mid g_t)$ computed over the recent past. As an example, because `grabing the cube' has more chance to lead to positive rewards for rare goals like `putting the cube higher', than other target goals such as `moving the gripper to the right', it will have a higher value. The next goal to target is finally sampled using a probability matching rule with an epsilon-greedy exploration:

$$
P_{select}(g_i) = \frac{\epsilon}{N} + (1 - \epsilon) \frac{value(g_t)}{\sum_i value(g_i)}
$$

\begin{figure}[ht]
\centering
    \includegraphics[width=1\columnwidth]{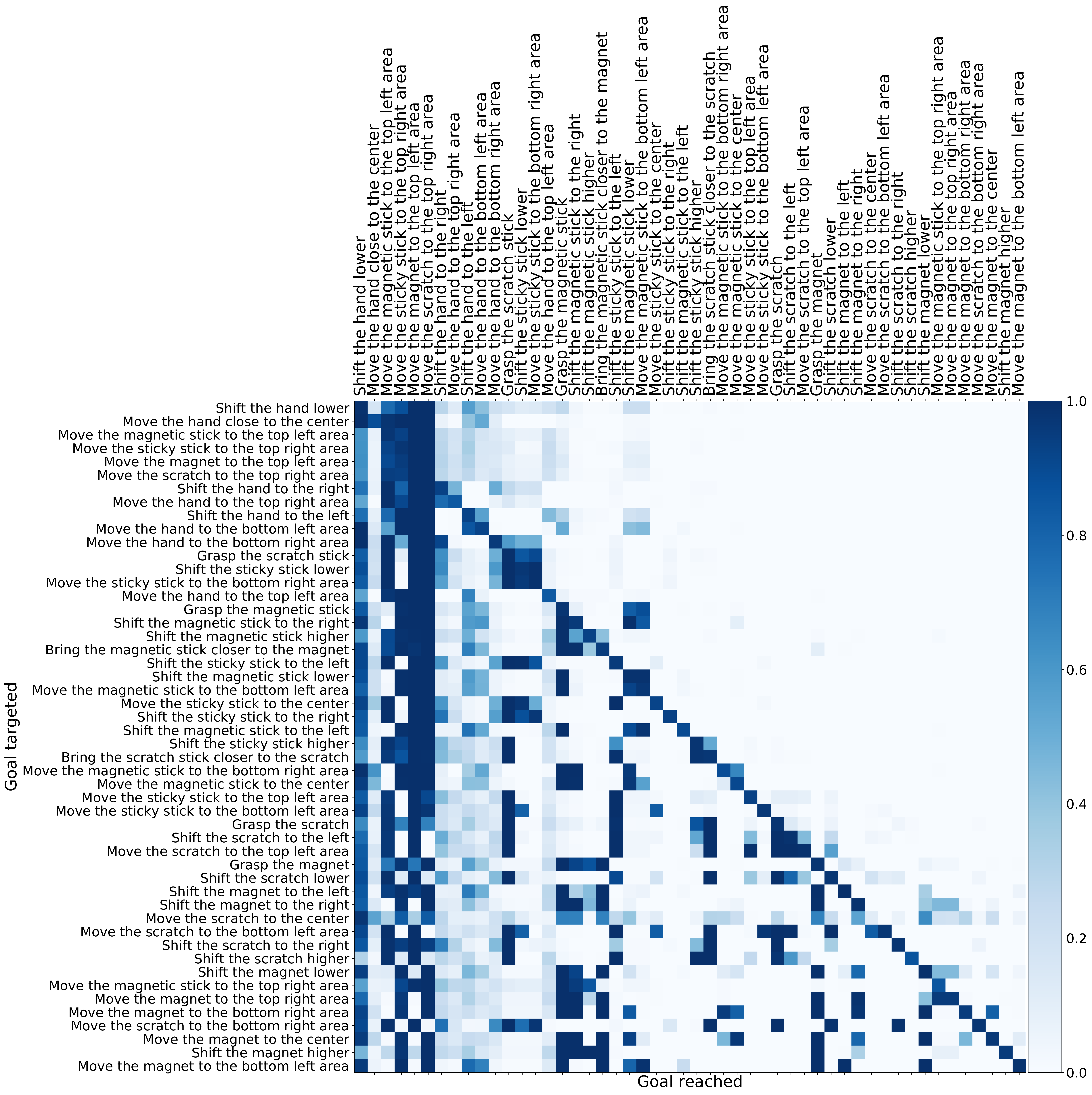}
     \caption{\textbf{Confusion matrix after learning.} This matrix depicts the frequency with which each of the goal is reached when each of the goal is targeted at the end of one particular run. Goals appear in the order of discovery. These statistics are tracked by the agent which uses it for its own curriculum learning.}  
\label{fig:confusion}
\end{figure}

Figure \ref{fig:confusion} shows the confusion matrix for goals at the end of learning for one particular run. For each of the potential goals that can be targeted, this matrix shows the frequency with which each of the goal is reached. This matrix is computed and updated internally by the agent as it learns to bias the sampling of goals to target towards goals leading to high-quality trajectories, i.e. trajectories that trigger SP feedback for rarely reached goals.

\paragraph{Intrinsic motivations for goal replay -} The second intrinsic motivation is used to bias the sampling of goals to substitute (replay) during learning. \ddpg is an off-policy algorithm, which means any transition could be used to learn about any goal. As in \unicorn and \curious \cite{mankowitz2018unicorn, colas2019curious}, we can use a transition collected while aiming at goal $g_i$ to learn about goal $g_j$ by simply substituting the goal encoding in the transition. Substituting goal encodings drives learning towards the substitute goals. Here we use a bias towards goals where the agent shows high absolute learning progress (LP). This mechanism was shown to generate an automatic curriculum in similar multi-goal settings \cite{colas2019curious}. The agent regularly performs self-evaluation rollouts without exploration noise and measures its competence (success rate on the recent past) and absolute learning progress on each of the goals as in \cite{colas2019curious}. The substitution probabilities of each goal are computed using a probability matching rule with epsilon-greedy exploration from the LP values ($\epsilon=0.2$).

\section{Learning a Goal-Conditioned Reward Function}
\label{SPfeedback}

The reward function is jointly learned with the policy. The agent collects data as the agent experiences the environment and use the descriptions provided by the SP to learn a model of a multi-goal reward function. 

\subsection{Data Collection} 

A challenge for learning the reward function is that the SP is only providing "positive" examples of what have been achieved: i.e. the agent is only collecting matching description of trajectories. In order to be able to learn a proper reward function, negative examples (i.e. example of trajectories and descriptions that do not match each other) are needed. As the SP is providing descriptions for each achieved goal during a trajectory, the agent can indirectly infer the goals that it has already discovered but were not achieved, and thus collect the missing negative examples. This way of collecting data is very convenient but leads to an imbalanced dataset.

\paragraph{Imbalance in terms of goals:} The agent is learning according to a natural curriculum that leads it to firstly attempt and master easy tasks before learning the harder ones. Thus, the distribution of the data collected also follows the same kind of curriculum and should be quite close to the distribution of state encountered by the agent. This leads to some goals being underrepresented compared to others: e.g. a newly discovered goal is necessarily underrepresented compared to older ones.

\paragraph{Imbalance between positive and negative rewards:}
It is obviously much more probable that the agent did not achieve a goal rather than it achieved it. This leads to collect much more negative examples than positive ones. For instance, when a goal has been achieved by chance once, it may take very long before it is reached again. Depending on the goals, the ratio between positive and negative examples can reach $95\%$ of negatives.

\subsection{Training} 
The problem of the reward function is framed as a goal-conditioned binary classification, meaning that the model has to learn a function $\mathcal{R}_i(s, \Delta s, g): \mathcal{S} \times \mathcal{S} \times \mathcal{G} \to \{ 0, 1\}$ where the goal $g$ is part of the input. Any classification model could be used but we choose a random forest as our preliminary test showed excellent performance and sample efficiency. Besides, it is fast to train. 

We want to ensure a good performance of the reward model uniformly on all the goals and not only on the ones that have been the most encountered. This requires to build an adapted train set to take into account the imbalance in the collected data. From the collected data, we sample when possible a similar proportion of data associated to each goals and ensure a minimal proportion positive examples of $20\%$. 

Once trained, the model can compute the reward associated to any transition ($s_t, \Delta_t, g$) to be used to update the goal-conditioned policy. (see subsection \ref{sec:algo_description} in the article).

\subsection{Monitoring}

It is all the more important that for newly discovered goals, the performance of the reward function gets good quickly as the agent could get stuck otherwise. We want to know in real time the performance of the reward function model to provide accurate reward to the model. By evaluating the learned reward function only on recently collected transitions and discovered goals, we can have an idea of how the helpfulness of the reward function model at present time. We monitor the performance of the reward function using the F1-score, precision and recall between the predicted rewards and the ground truth.

\begin{figure*}[ht!]
  \centering
  \subfigure[\label{fig:precision_avg}]{\includegraphics[width=0.45\textwidth]{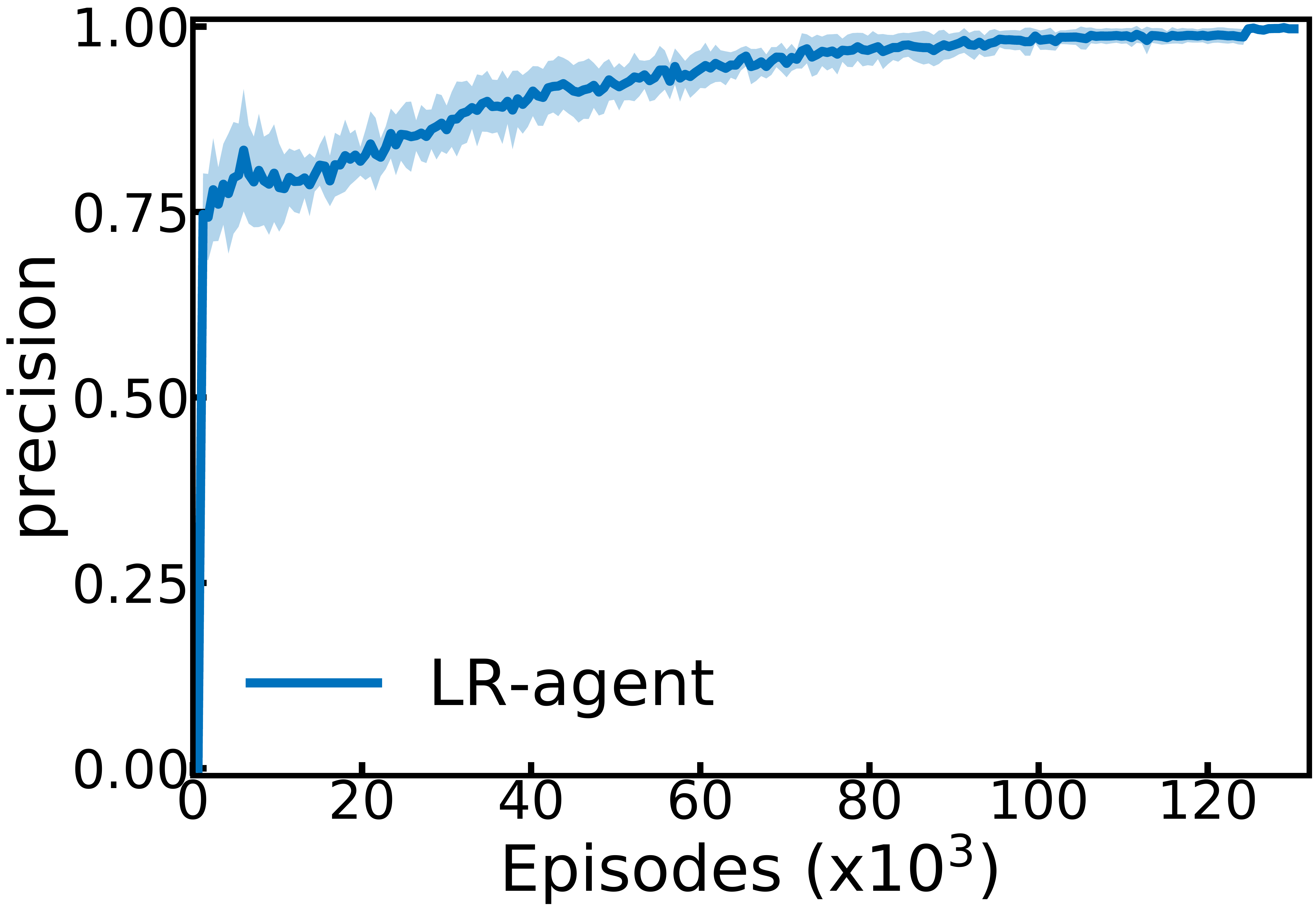}}
  \subfigure[\label{fig:recall_avg}]{\includegraphics[width=0.45\textwidth]{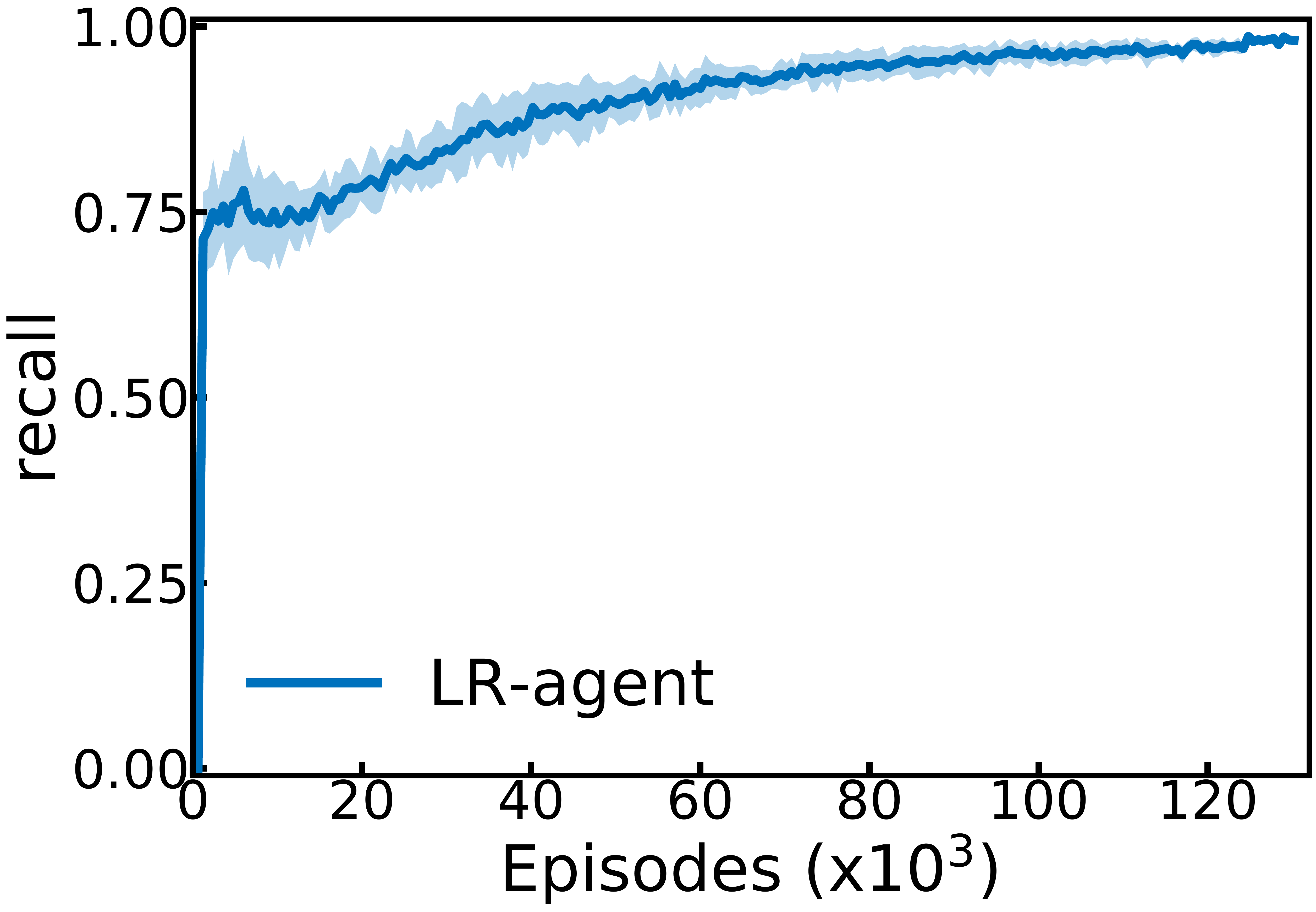}}
  \subfigure[\label{fig:precision_model}]{\includegraphics[width=0.45\textwidth]{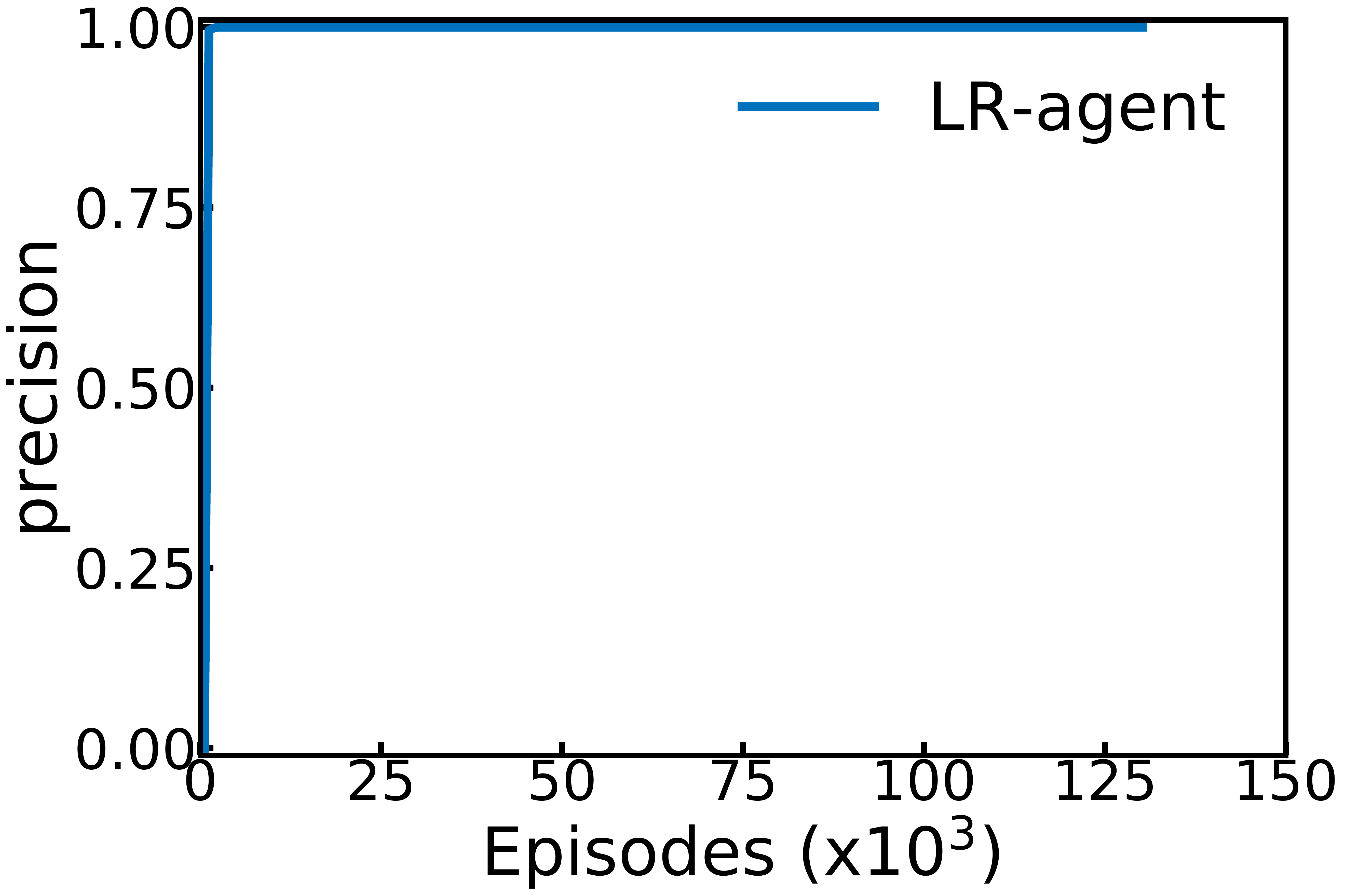}}
  \subfigure[\label{fig:recall_model}]{\includegraphics[width=0.45\textwidth]{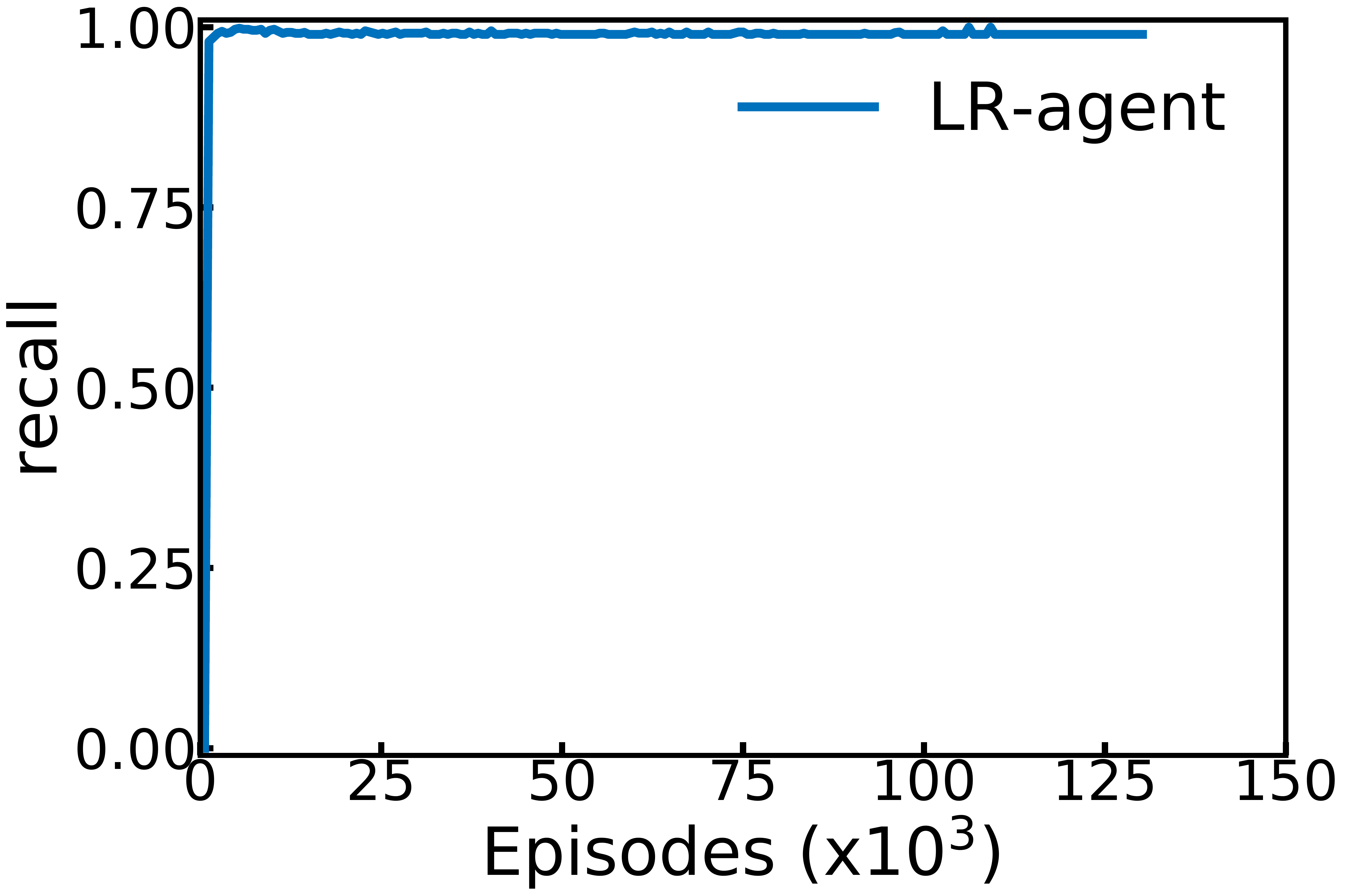}}
  \caption{\textbf{Reward function metrics.} a: Precision averaged over discovered goals over all 51 goals (mean, std). b: Recall averaged over discovered goals over all 51 goals (mean, std). c: Precision computed for the overall model (mean, std). d: Recall computed for the overall model (mean, std).}
\end{figure*}

The issue with a multi-goal reward function is that the metrics computed on a test set can be excellent, but hiding important discrepancies between the goals. The Figures \ref{fig:precision_avg} and \ref{fig:recall_avg} shows the evolution of the precision and recall computed for each goal and then averaged. The Figures \ref{fig:precision_model} and \ref{fig:recall_model} shows the evolution of the precision and recall computed for the whole model on a collected dataset containing the same proportion of each discovered goal. As we see, one could think by looking at the overall metrics, that the model gets almost perfect result from the beginning of the training. However, the quality of the reward model is very different from one goal to another and it needs around 80k episodes to be almost perfect.

\section{Environments Details}
\label{envdetails}
\begin{figure}[ht]
\centering
    \includegraphics[width=.5\textwidth]{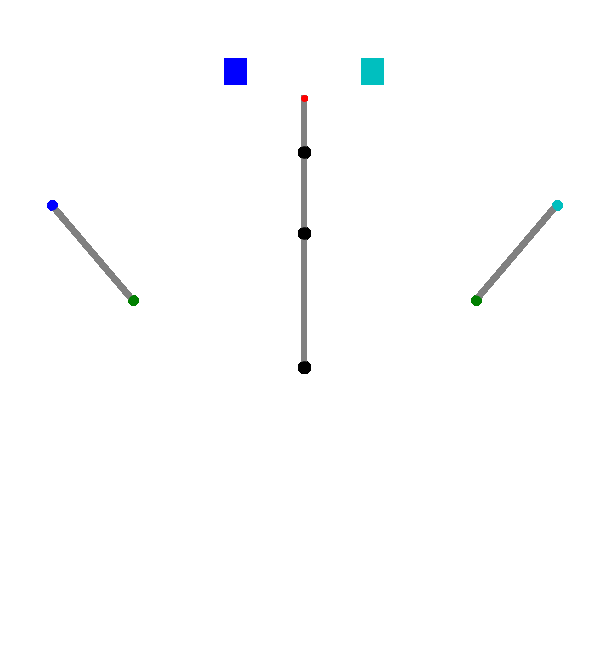}
        \caption{The \emph{ArmToolsToys} environment.}
\label{wrap-fig:1}
\end{figure} 
The agent is a 2D 3-joint robotic arm, controlled by the angular velocities of the joints and by an action on the gripper state (4D action space). Its environment is composed of two sticks and two objects. Each of the stick can be grabbed when the gripper is sufficiently close to the handle and when the gripper is closed. Each of the sticks enables to grasp a corresponding object. The agent receives as input the angular position of its joints, the 2D positions of its gripper, the handles, the end of the sticks and the objects (17D observation space). The social partner encodes 51 goals/descriptions in this environment:
\begin{itemize}
\footnotesize
    \item Description 0 : Shift the hand to the right
    \item Description 1 : Shift the hand to the left
    \item Description 2 : Shift the hand higher
    \item Description 3 : Shift the hand lower
    \item Description 4 : Move the hand close to the center
    \item Description 5 : Move the hand to the top right area
    \item Description 6 : Move the hand to the top left area
    \item Description 7 : Move the hand to the bottom right area
    \item Description 8 : Move the hand to the bottom left area
    \item Description 9 : Grasp the magnetic stick
    \item Description 10 : Grasp the scratch stick
    \item Description 11 : Shift the magnetic stick to the right
    \item Description 12 : Shift the magnetic stick to the left
    \item Description 13 : Shift the magnetic stick higher
    \item Description 14 : Shift the magnetic stick lower
    \item Description 15 : Move the magnetic stick to the center
    \item Description 16 : Move the magnetic stick to the top right area
    \item Description 17 : Move the magnetic stick to the top left area
    \item Description 18 : Move the magnetic stick to the bottom right area
    \item Description 19 : Move the magnetic stick to the bottom left area
    \item Description 20 : Shift the sticky stick to the right
    \item Description 21 : Shift the sticky stick to the left
    \item Description 22 : Shift the sticky stick higher
    \item Description 23 : Shift the sticky stick lower
    \item Description 24 : Move the sticky stick to the center
    \item Description 25 : Move the sticky stick to the top right area
    \item Description 26 : Move the sticky stick to the top left area
    \item Description 27 : Move the sticky stick to the bottom right area
    \item Description 28 : Move the sticky stick to the bottom left area
    \item Description 29 : Bring the magnetic stick closer to the magnet
    \item Description 30 : Bring the scratch stick closer to the scratch
    \item Description 31 : Grasp the magnet
    \item Description 32 : Grasp the scratch
    \item Description 33 : Shift the magnet to the right
    \item Description 34 : Shift the magnet to the left
    \item Description 35 : Shift the magnet higher
    \item Description 36 : Shift the magnet lower
    \item Description 37 : Move the magnet to the center
    \item Description 38 : Move the magnet to the top right area
    \item Description 39 : Move the magnet to the top left area
    \item Description 40 : Move the magnet to the bottom right area
    \item Description 41 : Move the magnet to the bottom left area
    \item Description 42 : Shift the scratch to the right
    \item Description 43 : Shift the scratch to the left
    \item Description 44 : Shift the scratch higher
    \item Description 45 : Shift the scratch lower
    \item Description 46 : Move the scratch to the center
    \item Description 47 : Move the scratch to the top right area
    \item Description 48 : Move the scratch to the top left area
    \item Description 49 : Move the scratch to the bottom right area
    \item Description 50 : Move the scratch to the bottom left area
\end{itemize}

\end{appendices}
\end{document}